\documentclass[english]{cccconf}
\usepackage[comma,numbers,square,sort&compress]{natbib}
\usepackage{multirow}
\begin{document}

\title{Spatial-Temporal Adaptive Graph Convolution with Attention Network for Traffic Forecasting}
\author{Weikang Chen\aref{scs}, 
        Yawen Li*\aref{sem},
        Zhe Xue\aref{scs},
        Ang Li\aref{scs},
        Guobin Wu\aref{didi}
        }


\affiliation[scs]{School of Computer Science (National Pilot School of Software Engineering), Beijing University of Posts and Telecommunications, Beijing Key Laboratory of Intelligent Telecommunication Software and Multimedia, P.~R.~China.
     }
\affiliation[sem]{School of Economics and Management, Beijing University of Posts and Telecommunications, P.~R.~China}
\affiliation[didi]{Didi Research Institute, Didi Chuxing, P.~R.~China.
        \email{weikangchen@bupt.edu.cn}}
\maketitle

\begin{abstract}
Traffic forecasting is one canonical example of spatial-temporal learning task in Intelligent Traffic System. Existing approaches capture spatial dependency with a pre-determined matrix in graph convolution neural operators. However, the explicit graph structure losses some hidden representations of relationships among nodes. Furthermore, traditional graph convolution neural operators cannot aggregate long-range nodes on the graph. To overcome these limits, we propose a novel network, Spatial-Temporal Adaptive graph convolution with Attention Network (STAAN) for traffic forecasting. Firstly, we adopt an adaptive dependency matrix instead of using a pre-defined matrix during GCN processing to infer the inter-dependencies among nodes. Secondly, we integrate PW-attention based on graph attention network which is designed for global dependency, and GCN as spatial block. What's more, a stacked dilated 1D convolution, with efficiency in long-term prediction, is adopted in our temporal block for capturing the different time series. We evaluate our STAAN on two real-world datasets, and experiments validate that our model outperforms state-of-the-art baselines.
\end{abstract}

\keywords{Traffic Forecasting, Adaptive Graph Convolution, Graph Attention Network, Spatial-Temporal Data}

\footnotetext{*Corresponding author: Yawen Li (warmly0716@126.com).}

\section{Introduction}

Traffic forecasting is regarded as one of the key components of Intelligent Traffic System. It aims to predict the future traffic flow trend based on the historical data, which is collected by the transportation departments. Recent studies show that graph convolution neural network can deal with this complex data structure - graph structure data - better than traditional convolution neural network. Spatial-temporal based on graph network models have been widely applied to solve these complex problems \cite{chen2021a,fang2021,Li2021,bai2020,Li14LPV}. While in traffic speed forecasting, the dataset is collected by the speed sensors every time from the city's road network. Each sensor is viewed as a node in the traffic graph network. The speed recorded by the sensors reflects road congestion. The higher speed sensors record, the less intense pressure traffic road withstands. What's more, each node in practice is always affected by its surrounding neighbors, which has been regarded as a basic assumption behind the spatial-temporal traffic forecasting problem. Imagine that if there was an accident on a road, especially at the crossroad, traffic paralysis would occur on adjacent sections of this road.

However, there are two problems when modeling the graph. On the one hand, most methods model the self-definition adjacent matrix based on Euclidean distance or Point of Interest as the relationship among the nodes. \cite{Wu2019,Li2017,Huang2020} process the adjacent matrix by road network distance with a thresholded Gaussian kernel. Firstly, the road status is changeable during the daytime. In general, traffic pressure is less at night and greater during the day in the city. Moreover, the traffic pressure on holidays is also different from that on weekdays. Secondly, the road network is three dimensions in our daily life. But when using the distance to calculate the weight matrix, it ignores the third dimension information. For example, although some roads on the overpass are close in space, the road conditions on each road do not affect the other roads. Inspired by the \cite{Bai2020}, we propose to enhance graph convolution network with the adaptive matrix to represent the hidden relationship among nodes.

\begin{figure}[!htb]
  \centering
  \includegraphics[width=\hsize]{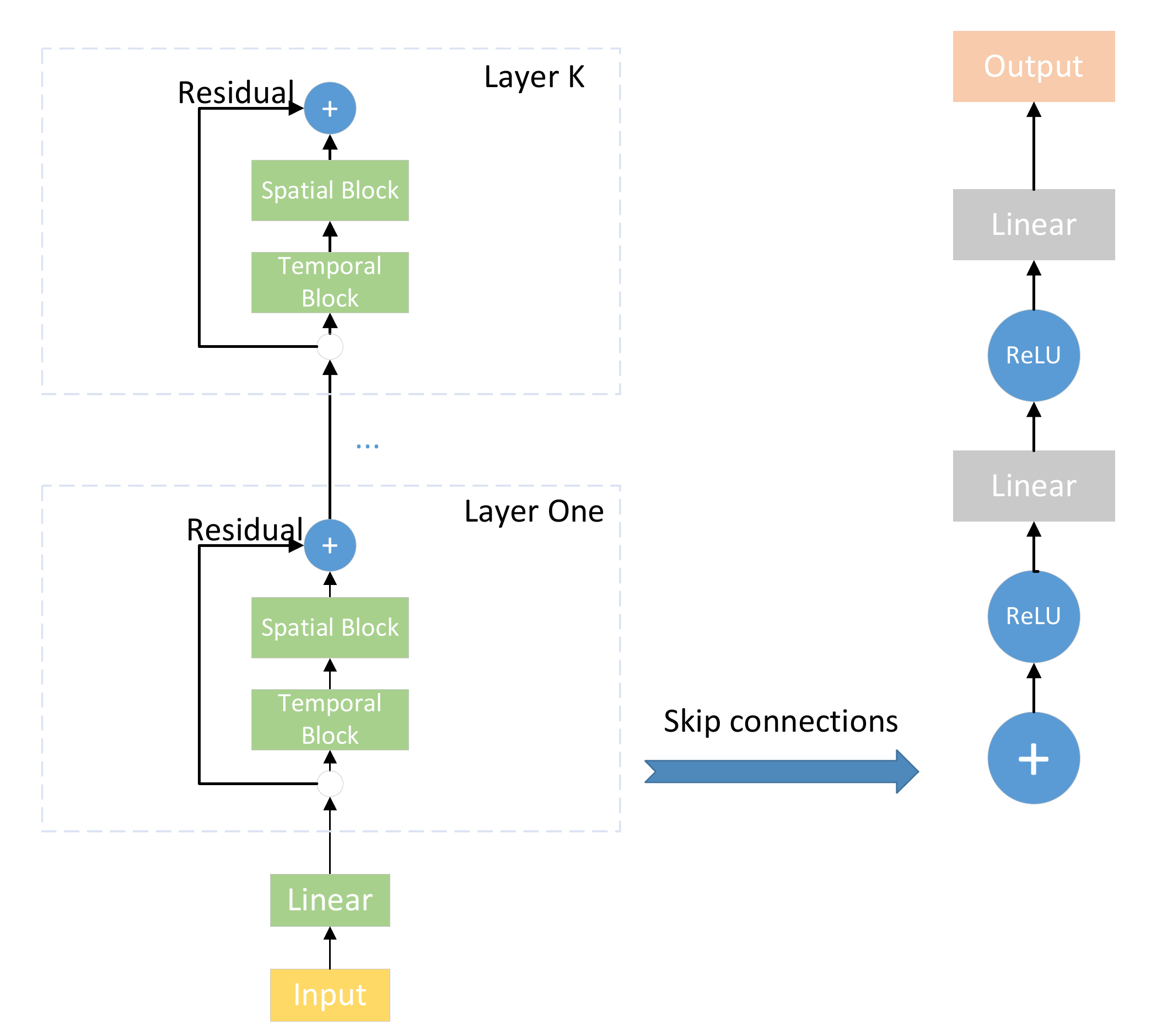}
  \caption{The Architecture of our model STAAN. On the left, it consists of K spatial-temporal blocks which use different time steps. With the skip connections of those different time steps outputs, there are two linear layers with activation layers on the right.}
  \label{fig1}
\end{figure}

On the other hand, graph convolution operator aggregates neighbors’ information by the adjacent matrix. For a long-term prediction, we need an operation to collect the global information among all nodes while the graph convolution operator prefers more on the local field. Attention mechanisms have become almost a standard in many sequence-based tasks \cite{Vaswani2017,Li13Region,Li17Distributed}. One of the benefits of attention mechanisms is that they focus on every relevant component to participle in making decisions. Graph attention network \cite{Velickovic2017,Ashish2017,Liang13Image} introduces an attention-based architecture to collect relevant information. With the efficiency of attention operation, graph attention network can be applied on graph nodes with specifying arbitrary weights. \cite{Park2020} uses spatial attention to capture the spatiotemporal dynamic in road networks. \cite{Huang2020} adopts the graph attention operation, which is called cosATT, to extract the similar road conditions of traffic networks. \cite{Huang2020,Yu2019} construct global and local extraction modules to consider the global information as well as local properties. So in our work, we propose PW-attention mechanism based on graph to capture the global similarity information among nodes to enhance the ability of graph representation with the global graph.

Convolutional operation processes a local neighborhood, either in time or space. \cite{Huang2020} adopt the gated mechanism to determine the ratio of information passed to the next layer. \cite{Wang2018} proposes a non-local operation that computes the response at a position as a weighted sum of the features at all positions in the input feature maps\cite{Li20Leveraging}. The set of positions can be in space, time or both of them. Traffic forecasting prediction is a long-term task. And non-local operations capture long-range dependencies directly by computing interactions between any two positions, regardless of their positional distance. With the assumption that similar roads have similar features, we combine the global graph attention as the similar function and graph convolution neural which is based on adaptive strategy, as the embedding function to enhance the ability to capture spatial features.

To accurately capture different time properties of traffic data, it is necessary to process the data through multiple time series scales. As for temporal prediction models, CNN-based approaches enjoy the advantages of parallel computing, stable gradients and low memory requirement \cite{Yan2018,bing2018} while RNN-based approaches suffer from time-consuming iterative propagation and gradient explosion/vanishing for capturing long-range sequences \cite{Defferrard2016}. However, to capture long sequences, these proposed models are much more complicated by using various 1D convolution kernel.

In our work, we design a novel architecture, the   Spatial-Temporal Adaptive graph convolution with Attention Network (STAAN) for graph-structured time series modeling in Figure 1, which addresses the two shortcomings we have aforementioned. Due to the achievement of \cite{Wu2019}, we adopt stacked dilated casual convolutions to capture temporal properties, which are based on CNN. Then motivated by the non-local, we combine the adaptive graph convolution neural with global graph attention neural to capture the local and global spatial properties. 

Our contributions are summarized as follows:
\begin{itemize}
\item We propose a novel deep learning framework named Spatial-Temporal Adaptive graph convolution network with Attention network (STAAN) for spatial-temporal traffic forecasting prediction by integrating PW-attention, based on graph attention network which is designed for global dependency, and adaptive GCN as spatial block.
\item In spatial blocks, we introduce an adaptive trainable weight matrix into graph convolution operator, instead of a pre-defined adjacent matrix and attention mechanism named PW-attention function which aims for global information.
\item We evaluate our STAAN on two real-world datasets, and experiments validate that our model outperforms state-of-the-art baselines.
\end{itemize}

\section{Related Works }

\subsection{Graph Neural Network}

Graph structured data is becoming more and more common in our daily life, like social connections, and traffic road networks. Motivated by the success of CNNs when dealing with lots of felids such as videos, natural language processing etc.\cite{Fang20Identity,Hu18Anomaly,Lin09Average,Li22Scientific}, researchers designed the graph convolution neural to deal with these graph-structured data \cite{Shi2019collaborative}. The methods of GCN are usually divided into two different types: spectral and spatial methods. The spectral method is using convolution definition to achieve in the graph. \cite{Thomas2017} introduced a Chebyshev polynomial parametrization for the spectral filter. \cite{Hamilton2017} provided a simplified version of ChebyNet, gaining success in the graph-based semi-supervised learning task. While spatial methods use a weighted average function by an adjacent matrix as the convolution operator. \cite{Velickovic2018} takes one-hop neighbors as neighborhoods and defines the weighting function as various aggregators over neighborhoods. Based on GCN, \cite{Li2017} combines graph convolution networks with recurrent neural networks in an encoder-decoder manner. \cite{bing2018} combines graph convolution with 1D convolution to tackle the time series prediction problem. However, during traffic time series, the graph-based dataset is changing over day and night, not staying in eternal status. Using the pre-defined adjacent matrix based on prior knowledge is not suitable for traffic prediction problems. To capture more information hidden in graph data, an adaptive matrix is a better ideal to extract properties during the processing of GCN.

Inspired by the attention mechanism, graph attention network [23] proposes to learn the weighting function via the self-attention mechanism which has been widely utilized in natural language processing, and image caption. Graph attention network performs self-attention on the nodes with a shared attentional mechanism $a$:
\begin{equation}
	\label{eq1}
 	{e_{ij}} = a(W[{h_i},{h_j}])
\end{equation}
where ${h_i} \in {R^C}$  and ${h_j} \in {R^C}$ are both the input features, $W \in {R^{C \times {C^{'}}}}$ is weight matrix, $a$ is a shared attention function. Then normalize the coefficient ${e_{ij}} \in R$:
\begin{equation}
	\label{eq2}
 	{\alpha _{ij}} = softmax_{j}({e_{ij}}) = \frac{{exp({e_{ij}})}}{{\sum\nolimits_{k \in {N_i}} {exp({e_{ik}})} }}
\end{equation}

Then the output representation is as follows:

\begin{equation}
	\label{eq3}
 	h_{i}^{\rm{'}} = \sigma (\sum\limits_{j \in {N_i}} {{\alpha _{ij}}W{h_i})} 
\end{equation}

where ${N_i}$ is the set of neighbors of node i and $\sigma ( \cdot )$ is an activation function. Recently, \cite{Lingbo2018,Shengnan2019} propose a novel multi-level attention-based network. \cite{Zheng2020} propose ASTGCN and \cite{Jiani2018} propose GMAN. \cite{Park2020} uses spatial attention to capture the spatiotemporal dynamic in road networks. \cite{Huang2020} adopts the graph attention operation to extract the similar road conditions of traffic networks. Different from these works, our works adopt a dynamic trainable parameter matrix and attention mechanism as a similar measurement to automatically reflect and capture the hidden connection among all nodes.

\subsection{Spatial-Temporal Prediction }
In recent years, researchers have conducted on traffic forecasting problems. It has been viewed as a spatial-temporal forecasting task. However, there are two ways to capture the spatial-temporal. One is learning features with two components to capture temporal and spatial separately. \cite{Wu2019} introduces a self-adaptive graph and dilated convolution to capture spatial and temporal dependency.\cite{Park2020} is an end-to-end solution for traffic forecasting that captures spatial, short-term, and long-term periodical dependencies. \cite{Bing2018} uses spatial attention, temporal attention, and spatial sentinel vectors to capture the spatiotemporal dynamic in road networks. The other is to model spatial-temporal correlations simultaneously. \cite{Guo2019} combines graph convolution with 1D convolution. And \cite{Zhang2020} introduces 3D convolutions into this area, which can effectively extract features from both the spatial and temporal dimensions. \cite{Geng2019} uses Pseudo three-dimensional convolution network to combine structure learning convolution, which is designed to learn graph structure, with the time feature. Due to capturing heterogeneity of spatial and temporal properties simultaneously, those methods need to use more layers with different convolution operators. In all, to be computationally efficient heterogeneity between the spatial feature and temporal feature, our model includes two components to capture the spatial and temporal feature in the traffic data separately.

\section{Methodology}
In this section, we first give the mathematical definition of the traffic prediction and then show the details about our framework. 

\subsection{Traffic Prediction Problem}
We target on the multi-step traffic forecasting problem. Consider multitudinous traffic series that contains history time series $X = \{ {X_{:,0}},{X_{:,1}},...,{X_{:,t}}\} $, and for each time step,${X_{:,t}} = {\{ {x_{1,t}},{x_{2,t}},...,{x_{i,t}}...,{x_{N,t}}\} ^T} \in {R^{N \times C}}$ , where N is the number of all nodes and C denotes the features of each node. Our target is to predict the future features of all nodes by using the history data X. 
\begin{equation}
	\label{eq4}
 	[{X_{:,0}},{X_{:,1}},...,{X_{:,t}},G] \to F[{X_{:,t + 1}},...,{X_{:,t + T}}]
\end{equation}
  
where $G = (V,E,A)$ , which denotes the road network. And $V$  is a set of N nodes; $E$ is a set of edges; $A \in {R^{N \times N}}$ corresponds to the adjacency matrix. 

\begin{figure}[!htb]
  \centering
  \includegraphics[width=0.6\hsize]{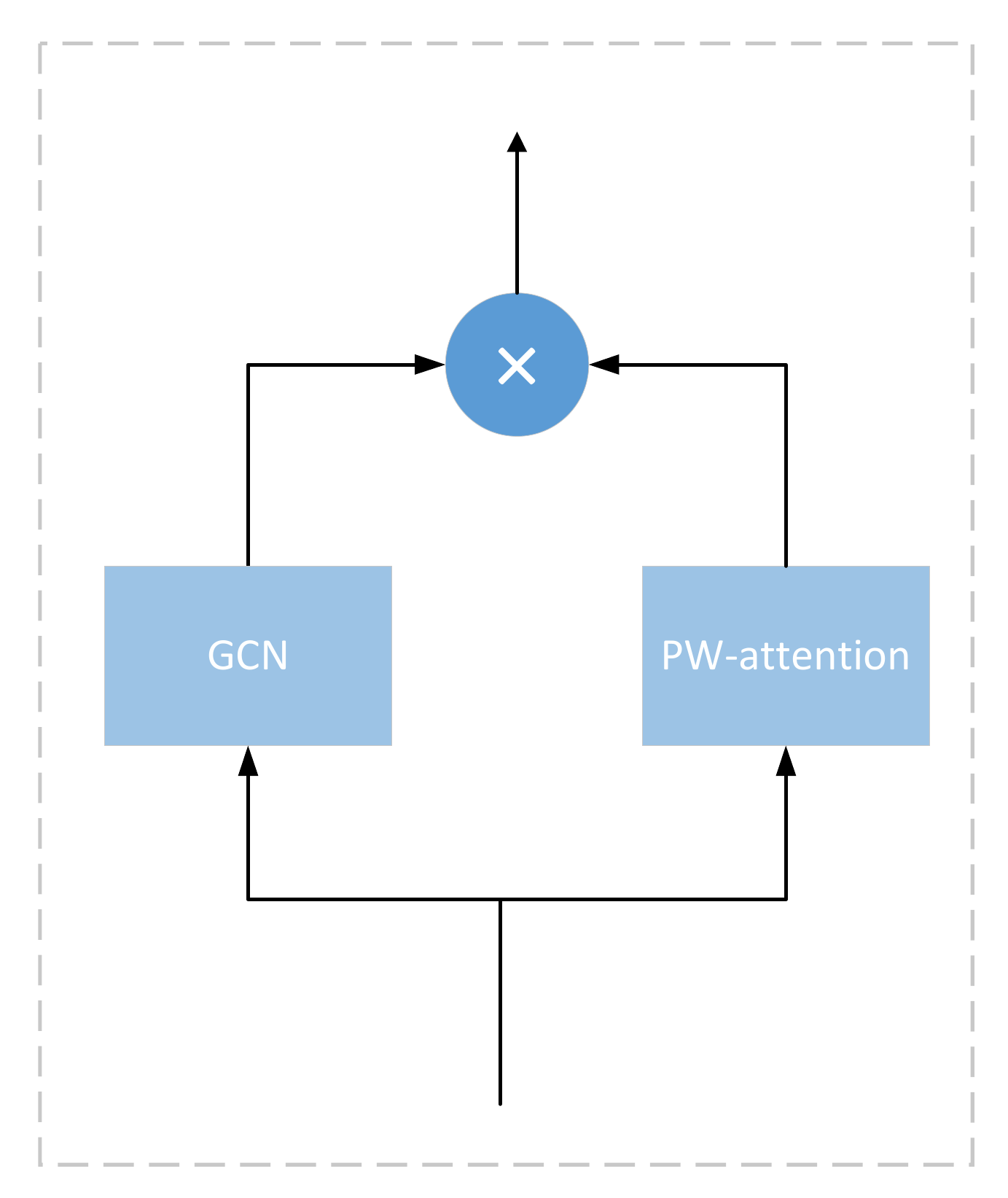}
  \caption{Spatial Block. It contains of GCN and PW-attention components which captures the local and global spatial data.}
  \label{fig2}
\end{figure}

\subsection{Spatial Block with Graph Neural Network}

\textbf{Adaptive Matrix:} Graph convolution is an essential operation to learn the nodes features. Most recent work in traffic forecasting deploys GCN to capture the spatial features. According to [21], the graph convolution operation can be well-approximated by 1st order Chebyshev polynomial expansion and generalized to high-dimensional GCN as:
\begin{equation}
\label{eq5}
\\Z = ({I_N} + {D^{ - \frac{1}{2}}}A{D^{ - \frac{1}{2}}})XW
\end{equation}

where $A \in {R^{N \times N}}$ is the normalized adjacency matrix, $D$ is the degree matrix, and $I_{N}$ is the identity matrix of n dimensions. And $X \in {R^{N \times C}}$ denote the input channels,$W \in {R^{C \times F}}$ is the layer-specific trainable weight matrix. GCN operator can divide into two operations: (1) aggregate the features from its neighbors. (2) transform the features to a high dimensional space with the weight matrix. So in the first step, the effect of aggregation will be influenced by the adjacent matrix. Existing works mainly utilize the pre-defined adjacent matrix for GCN operation. \cite{Li2017} use the distance to compute the pairwise road network and build the adjacency matrix using the thresholded Gaussian kernel. \cite{Yu2015} uses the similarity of the features to create the adjacent matrix. However, these calculated manually the matriax cannot truly reflect the hidden connections between the nodes. \cite{Wu2019,Bai2020}proposes an adaptive matrix by multiplying the embedding of nodes ${E_A} \in {R^{N \times d}}$ and ${E_A}^T$ to replace the static adjacent matrix:
\begin{equation}
	\label{eq6}
	\\{A_{adaptive}} = g({E_A} \bullet {E_A}^T)
\end{equation}

where $g( \cdot )$  is a function to normalize the dynamic matrix. Instead of computing the degree matrix repeatedly, we can use the trainable ${E_A}$ during the data training. Thus, the GCN can be formulated as:
\begin{equation}
	\label{eq7}
	\\Z = ({I_N} + {A_{adaptive}})XW
\end{equation}

First, a dynamic trainable weight matrix, obtained by the dot product of the embedding nodes and its transposition, can truly reflect the hidden connections instead of using the pre-defined matrix. Rather than using the trainable parameter $A \in {R^{N \times N}}$ , embedding function ${E_A} \in {R^{N \times d}}$ reduces the space complexity for faster training when $d \ll N$.

\textbf{Graph Attention:} A self-attention \cite{Vaswani2017} module computes the response at a position in a sequence by attending to all positions and taking their weighted average in an embedding space. Graph attention networks \cite{Defferrard2016} is to compute the hidden representations of each node in the graph, following the self-attention strategy. With the assumption that similar roads have similar features. Like \cite{Huang2020}, we use the global graph attention networks to learn the similarity of the nodes instead of neighbors in graph attention networks. We define PW-attention operator:
\begin{equation}
\label{eq8}
{a_{ij}} = sigmoid(similarity({w_{ij}}[{x_i},{x_j}]))
\end{equation}
\begin{equation}
\label{eq9}
output_i = \sum\limits_{j \in {N_i}} {{a_{ij}}{w_{ij}}{x_j}}
\end{equation}
where $similarity( \cdot )$ is a pairwise function compute a scalar which representing relationship between node i and j. And ${N_i}$ is the all nodes. Different from GAT, we use the global graph to obtain relationship among nodes to enhance the ability to capture similar nodes’ features.

\textbf{Spatial Block:} In our work, we use the GCN as the representation function and GAT as a similar function to learn the spatial features during the dataset. As shown in Figure 2, the spatial block work as follows. First, we pass the input $H \in {R^{N \times C}}$ into GCN to capture the neighbors’ feature, where $N$ denotes the number of nodes, and $C$ is the embedding dimension. On the other hand, we use the global graph PW-attention to learn the similarity among different nodes. Then use the element-wise Hadamard product on the above output. Experiments show that our spatial block performs well with the special module, which combines adaptive graph convolution network with PW-attention mechanism.

\subsection{Temporal Block with Temporal Convolution Neural}

We adopt the dilated causal convolution \cite{Shi2020} as our temporal convolution neural layer to capture the temporal features trends. Unlike RNN-based methods, models with causal convolutions can train faster to obtain the long-range time series because of the non-recurrent connections. Figure 3 depicts dilated causal convolutions for dilations 1, 2, and 4. At the beginning of dilated causal convolution, it preserves the temporal causal order by padding zeros to the inputs. On the first layer, the dilated causal convolution operation slides over inputs by skipping values with the 1 step. And on the second layer, the step adds up to 2, which means the convolution just keeps some necessary information for the next layer. After stacked dilated convolutions, we will get a greatly large receptive field with a few layers. We adopt the same gated mechanisms in the dilated convolution in Figure 4, which has been proved to control the features. For each layer:
\begin{equation}
	\label{eq10}
	z = g({W_{f,k}}x) \otimes \sigma ({W_{g,k}}x)
\end{equation}

where $ \otimes $  denotes an element-wise product operator, function $g( \cdot )$  is an activation function of the output and $\sigma ( \cdot )$ is the sigmoid function which decides whether the information should forget. K is the layer index, f and g denote the filter and gate. And W is a learnable parameter. We set the $g( \cdot )$  as tangent hyperbolic function.

\begin{figure}[!htb]
  \centering
  \includegraphics[width=\hsize]{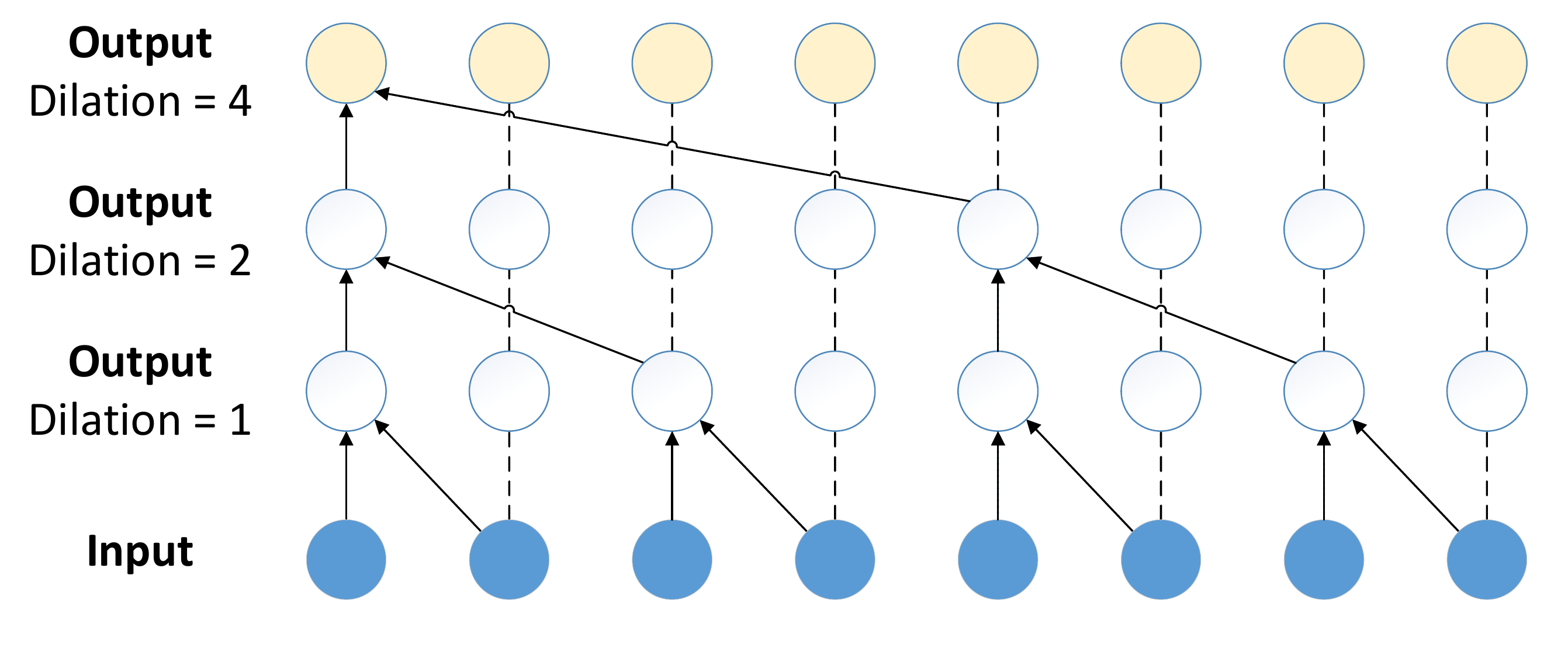}
  \caption{A stack of dilated casual convolution layers.}
  \label{fig3}
\end{figure}

\subsection{Architecture of Network }

As shown in Figure 1, our framework of STAAN consists of stacked spatial-temporal blocks with skip connections to predict future traffic forecasting results. Each spatial-temporal block includes the temporal convolution neural (capture the time features) and the graph convolution networks (capture the space features). Our model can handle different temporal levels with different spatial-temporal blocks. For example, at the bottom block of our model, it can deal with the shortest-term time series, while from the bottom to the top, our model can receive more long-term time information. To keep the short-term information, we adopt a cascaded connection the higher-level uses the lower level’s output as its input. So when the input of each block is a tensor H with the size of [C, N, L] where C denotes embedding dimension, N is the number of all nodes, and L is the length sequence we want to predict. Spatial block is applied to each $h[:,:,i] \in {R^{N \times C}}$ to capture the spatial information. In the end, a skip connection is a great way to collect all the output from every block to predict the results.

\subsection{Loss Function}

We use the mean absolute error (MAE) to measure the performance of our model which is formulated as follows:
\begin{equation}
	\label{eq11}
	L({\hat{X}_{j:t+1}},...,\hat{X}_{j:t+T};\Theta) = \frac{1}{n}\sum\limits_{i = 1}^T {\sum\limits_{j = 1}^{N} {|{\hat{X}_{j:i}} - {X}_{j:i}|} } 
\end{equation}
where $X = \{ {X_{:,0}},{X_{:,1}},...,{X_{:,t}}\} $.

\begin{figure}[!htb]
  \centering
  \includegraphics[width=0.6\hsize]{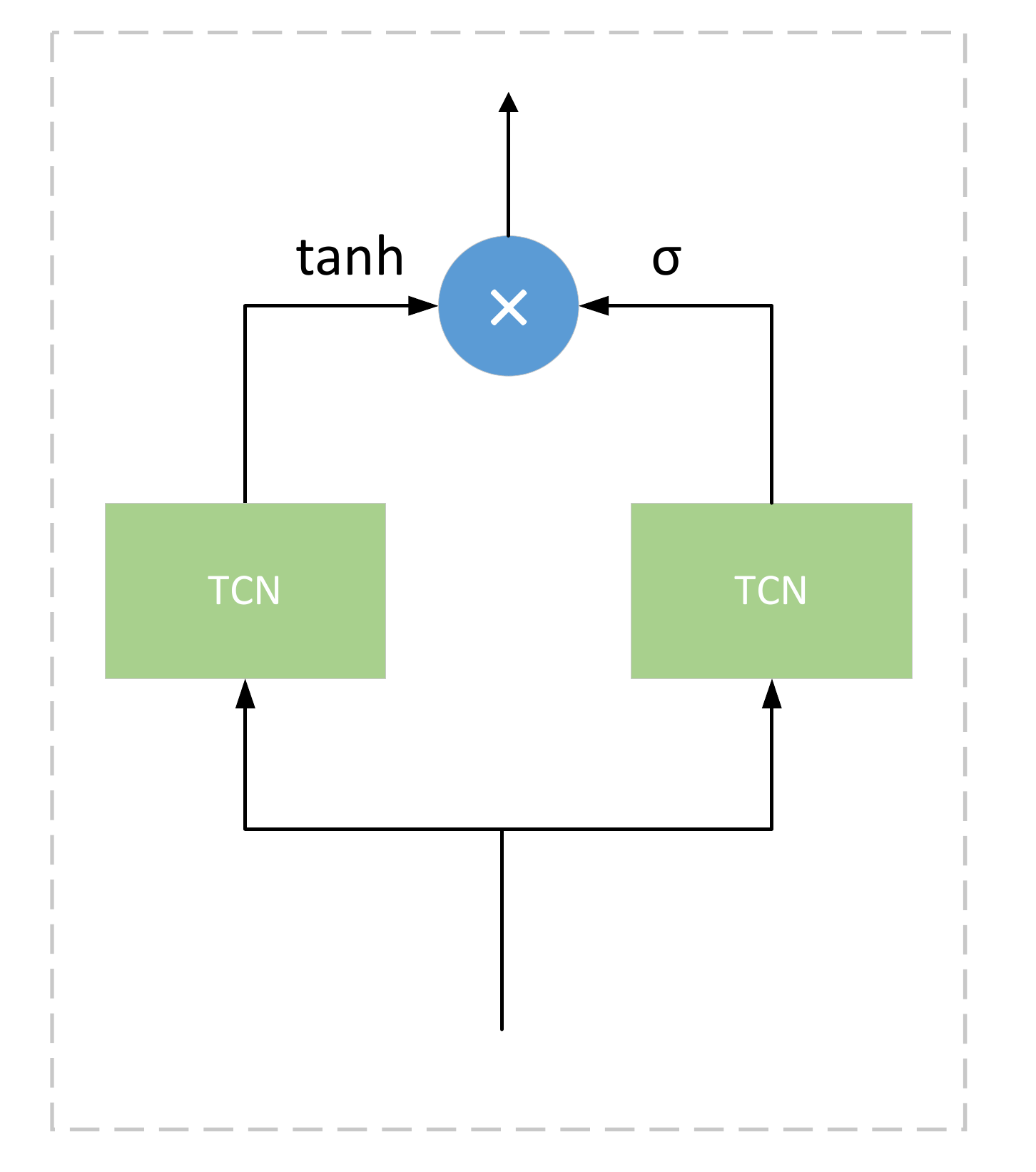}
  \caption{Temporal Block.It contains dilated casual convolution layer with gated mechanisms.}
  \label{fig4}
\end{figure}

\begin{table*}[]
  \centering
  \caption{Performance comparison of STAAN and other baseline models. STAAN achieves the best results on both datasets}
  \label{tab1}
\begin{tabular}{|c|ccc|ccc|}
\hline
\multirow{2}{*}{Model} & \multicolumn{3}{c|}{METR-LA(15/30/60 min)}                                                                                   & \multicolumn{3}{c|}{PEMS-BAY(15/30/60 min)}                                                                                       \\ \cline{2-7} 
                       & \multicolumn{1}{c|}{MAE}                       & \multicolumn{1}{c|}{MAPE(\%)}                  & RMSE                       & \multicolumn{1}{c|}{MAE}                         & \multicolumn{1}{c|}{MAPE(\%)}                    & RMSE                        \\ \hline
ARIMA                  & \multicolumn{1}{c|}{3.99/ 5.15/ 6.90}          & \multicolumn{1}{c|}{9.60/ 12.70/ 17.40}        & 8.21/ 10.45/ 13.23         & \multicolumn{1}{c|}{1.62/ 2.33/ 3.38}            & \multicolumn{1}{c|}{1.62/ 2.33/ 3.38}            & 1.62/ 2.33/ 3.38            \\ \hline
FC-LSTM                & \multicolumn{1}{c|}{3.44/ 3.77/ 4.37}          & \multicolumn{1}{c|}{9.60/ 10.90/ 13.20}        & 6.30/ 7.23/ 8.69           & \multicolumn{1}{c|}{2.05/ 2.20/ 2.37}            & \multicolumn{1}{c|}{4.80/ 5.20/ 5.70}            & 4.19/ 4.55/ 4.69            \\ \hline
DCRNN                  & \multicolumn{1}{c|}{2.77/ 3.15/ 3.60}          & \multicolumn{1}{c|}{7.30/ 8.80/ 10.50}         & 5.38/ 6.45/ 7.59           & \multicolumn{1}{c|}{1.38/ 1.74/ 2.07}            & \multicolumn{1}{c|}{2.90/ 3.90/ 4.90}            & 2.95/ 3.97/ 4.74            \\ \hline
GGRU                   & \multicolumn{1}{c|}{2.71/ 3.12/ 3.64}          & \multicolumn{1}{c|}{6.99/ 8.56/ 10.62}         & 5.24/ 6.36/ 7.65           & \multicolumn{1}{c|}{-}                           & \multicolumn{1}{c|}{-}                           & -                           \\ \hline
STGCN                  & \multicolumn{1}{c|}{2.87/ 3.48/ 4.45}          & \multicolumn{1}{c|}{7.40/ 9.40/ 11.80}         & 5.54/ 6.84/ 8.41           & \multicolumn{1}{c|}{1.46/ 2.00/ 2.67}            & \multicolumn{1}{c|}{3.01/ 4.31/ 5.73}            & 2.90/ 4.10/ 5.40            \\ \hline
APTN                   & \multicolumn{1}{c|}{2.76/ 3.15/ 3.70}          & \multicolumn{1}{c|}{7.30/ 8.80/ 10.69}         & 5.38/ 6.43/ 7.69           & \multicolumn{1}{c|}{1.38/ 1.97/ 2.33}            & \multicolumn{1}{c|}{2.91/3.69/4.65}              & 2.96/ 3.95/ 4.60            \\ \hline
ST-UNet                & \multicolumn{1}{c|}{2.72/ 3.12/ 3.55}          & \multicolumn{1}{c|}{6.90/ 8.40/ 10.00}         & 5.13/ 6.16/ 7.40           & \multicolumn{1}{c|}{2.15/ 2.81/ 3.38}            & \multicolumn{1}{c|}{4.03/ 5.42/ 6.68}            & 5.06/ 6.79/ 8.33            \\ \hline
GWN                    & \multicolumn{1}{c|}{2.69/ 3.07/ 3.53}          & \multicolumn{1}{c|}{6.90/ 8.37/ 10.01}         & 5.15/ 6.22/ 7.37           & \multicolumn{1}{c|}{1.30/ 1.63/ 1.95}            & \multicolumn{1}{c|}{2.70/ 3.70/ 4.60}            & 2.74/ 3.70/ 4.52            \\ \hline
ST-GRAT                & \multicolumn{1}{c|}{2.60/ 3.01/ 3.49}          & \multicolumn{1}{c|}{6.61/8.15/ 10.01}          & \textbf{5.07}/ 6.21/ 7.42  & \multicolumn{1}{c|}{\textbf{1.29/   1.61/ 1.95}} & \multicolumn{1}{c|}{\textbf{2.67/   3.63/}4.64} & \textbf{2.71/   3.69/ 4.54} \\ \hline
SLCNN                  & \multicolumn{1}{c|}{\textbf{2.53/ 2.88/ 3.30}} & \multicolumn{1}{c|}{\textbf{5.18/ 6.15/ 7.20}} & 5.18/ \textbf{6.15/ 7.20}  & \multicolumn{1}{c|}{1.44/ 1.72/ 2.03}            & \multicolumn{1}{c|}{2.90/ 3.81/ \textbf{4.53}}   & 3.00/ 3.90/ 4.80   \\ \hline
STAAN                  & \multicolumn{1}{c|}{\textbf{2.49/ 2.85/ 3.29}} & \multicolumn{1}{c|}{\textbf{5.15/ 6.13/} 7.23} & \textbf{5.10/ 6.10 / 7.15} & \multicolumn{1}{c|}{\textbf{1.27/ 1.57/   1.92}} & \multicolumn{1}{c|}{\textbf{2.67/} 3.65/ \textbf{4.51}}   & \textbf{2.58/ 3.57/ 4.40}   \\ \hline
\end{tabular}
\end{table*}

\section{Experiments}

\subsection{Dataset}

In the experiment, we use the two real-world traffic datasets, namely METR-LA and PEMS-BAY released by \cite{Li2017}. More details for the datasets are in Table \ref{tab2}.

\textbf{METR-LA.} Traffic speed prediction on the METR-LA dataset, which contains 4 months of data recorded by 207 loop detectors ranging from March 1, 2012, to June 30, 2012, on the highway of Los Angeles.

\textbf{PEMS-BAY.} Traffic speed prediction on the PEMS-BAY dataset, which contains 6 months of data recorded by 325 sensors ranging from January 1, 2017, to June 30, 2017, in the Bay Area.

\subsection{Baselines}
We adopt Mean Absolute Errors(MAE), Mean Absolute Percentage Errors (MAPE), and Root Mean Squared Errors (RMSE) to compare the performance STAAN with the following state-of-the-art models:
\begin{itemize}
\item ARIMA. Auto-Regressive Integrated Moving Average model with Kalman filter \cite{Wu2019}.
\item FC-LSTM. Recurrent neural network with fully connected LSTM hidden units \cite{Wu2019}.
\item DCRNN. Diffusion convolution recurrent neural network \cite{Wu2019}, which combines graph convolution networks with recurrent neural networks in an encoder-decoder manner.
\item GGRU. Graph gated recurrent unit network \cite{Defferrard2016}. Recurrent-based approaches. GGRU uses attention mechanisms in graph convolution.
\item STGCN. Spatial-temporal graph convolution network \cite{bing2018}, which combines graph convolution with 1D convolution.
\item GWN. A convolution network architecture \cite{Wu2019}, introduces a self-adaptive graph to capture the hidden spatial dependency, and uses dilated convolution to capture the temporal dependency.
\item APTN. Attention-based Periodic-Temporal neural Network \cite{Park2020}, which is an end-to-end solution for traffic forecasting that captures spatial, short-term, and long-term periodical dependencies.
\item ST-UNet. Spatial-Temporal U-Net \cite{Yu2019} adopts a U-shaped network to extract temporal and spatial properties simultaneously.
\item ST-GRAT. Spatial-Temporal Graph attention \cite{Bing2018}, which uses spatial attention, temporal attention, and spatial sentinel vectors to capture the spatiotemporal dynamic in road networks.
\item SLCNN. Spatial-temporal Graph Structure Learning \cite{Geng2019} proposes pseudo three-dimensional convolution, which combines with the structure learning convolution to capture the temporal dependencies in traffic data.
\end{itemize}

\subsection{Experimental Settings}

All experiments are performed on a Linux server with one Intel(R) Xeon(R) Gold 5218 CPU @ 2.30GHz and one NVIDIA GeForce RTX 2080 Ti CPU card. We use the eight blocks for capture the time sequence. And in temporal convolution neural, we use the sequence 1, 2, 1, 2, 1, 2, 1, 2 as the dilation sequence for each block. In spatial block, we use the softmax function as normalization in Equation ~(\ref{eq1}). During train model, we use the Adam optimizer as our optimizer, with initial value 0.001 and we adopt dropout with 0.3 for graph neural network. And we set the input dimension to 2, number of hidden layer to 40 and the output length sequence to 12. For dataset, we split it to 7:2:1 as the training data, valid data and the test data. All tests adopt 5 minutes as the time windows. And the target we want to get from our model is the future time in 15, 30, 60 minutes.

\subsection{Experimental Results}

\begin{table}[!htb]
  \centering
  \caption{The details for the datasets}
  \label{tab2}
  \begin{tabular}{|c|cc|}
  \hline
              & \multicolumn{1}{c|}{METR-LA}           & PEMS-BAY          \\ \hline
Nodes         & \multicolumn{1}{c|}{207}               & 325               \\ \hline
Records       & \multicolumn{1}{c|}{23974}             & 52093             \\ \hline
Time span     & \multicolumn{1}{c|}{2012.03 - 2012.06} & 2017.01 - 2017.06 \\ \hline
Time interval & \multicolumn{2}{c|}{5 minutes}                             \\ \hline
Daily range   & \multicolumn{2}{c|}{0:00-24:00}                            \\ \hline
	\end{tabular}
\end{table}

Table \ref{tab1} shows the performance among these models for 15, 30, and 60 minutes ahead prediction on METR-LA and PEMS-BAY datasets. STANN gets the greatest performance in our experiment in particular.
Several observations can be get by following analyses. Firstly, STAAN achieves a small improvement than SLCNN and GWN. We think our architecture is more capable of detecting spatial properties with PW-attention for global information instead of structure learning. Secondly, ST-GRAT uses spatial attention and temporal attention to capture the spatiotemporal properties. In contrast, STAAN employs stacked spatial-temporal blocks with different time steps for GCN blocks with different parameters. Thirdly, APTN, ST-GRAT have better results than STGCN, GGRU since APTN and ST-GRAT adopt attention mechanisms for both spatial and temporal properties. And ST-UNet adopts U-Net to capture spatial and temporal properties simultaneously, it performs the same as DCRNN at the short term prediction but shows better in the long term. FC-LSTM based on fully connected LSTM hidden units obtains a better than the traditional method. While statistical method ARIMA has poor performance since its ability cannot deal with heterogeneous data.


\section{Conclusion}

In this paper, we present a spatial-temporal adaptive graph convolution with the attention mechanism network, STAAN for short, to address the traffic forecasting problem. STAAN composes spatial blocks based on GCN and temporal blocks based on TCN to capture heterogeneous graph structure data. Furthermore, STAAN utilizes an adaptive matrix and PW-attention mechanism which aims for learning global information. Validated on two real-world datasets, STAAN demonstrates the best performance than baselines for traffic forecasting. In future work, we will focus on large heterogeneous datasets with more novel graph convolution neural network.

\end{document}